\documentclass[10pt,twocolumn,letterpaper]{article}
\usepackage[table,x11names]{xcolor}
\usepackage{iccv}
\usepackage{times}
\usepackage{epsfig}
\usepackage{graphicx}
\usepackage{amsmath}
\usepackage{amssymb}\usepackage{multicol}
\usepackage{multirow}
\usepackage[T1]{fontenc}
\usepackage[latin9]{inputenc}
\usepackage[english]{babel}
\usepackage{collcell}
\usepackage{hhline}
\usepackage{pgf}
\usepackage{multirow}
\usepackage{booktabs}

\usepackage{collcell}
\usepackage{array}
\usepackage{tikz}
\usepackage{pgfkeys}
\usepackage{graphicx}

\usepackage[pagebackref=true,breaklinks=true,letterpaper=true,colorlinks,bookmarks=false]{hyperref}

\usepackage{multicol}
\usepackage{multirow}

\usepackage{authblk}
\usepackage{colortbl}
\definecolor{Pblue}{RGB}{0,140,204}
\usepackage{pgfplotstable}
\usepgfplotslibrary{colormaps}

\makeatletter
\renewcommand*{\@fnsymbol}[1]{\ensuremath{\ifcase#1\or \sharp \or * \or \dagger\or
    \mathsection\or \mathparagraph\or \|\or **\or \dagger\dagger
    \or \ddagger\ddagger \else\@ctrerr\fi}}
\makeatother

\def\cca#1{%
    \pgfmathsetmacro\calct{min(max((#1-25)*100/(48-25),0),100)}%
    \pgfmathsetmacro\calcs{min(max((#1-2)*100/(25-2),0),100)}%
    \edef\clrmacrot{\noexpand\cellcolor{OrangeRed1!\calct!Gold1}}%
    \edef\clrmacros{\noexpand\cellcolor{Gold1!\calcs!SpringGreen4}}%
    \clrmacros%
    \ifdim #1 pt>25pt \clrmacrot\fi{#1}%
}

\def\ccb#1{%
    \pgfmathsetmacro\calct{min(max((#1-15)*100/(25-15),0),100)}%
    \pgfmathsetmacro\calcs{min(max((#1-4)*100/(15-4),0),100)}%
    \edef\clrmacrot{\noexpand\cellcolor{OrangeRed1!\calct!Gold1}}%
    \edef\clrmacros{\noexpand\cellcolor{Gold1!\calcs!SpringGreen4}}%
    \clrmacros%
    \ifdim #1 pt>15pt \clrmacrot\fi{#1}%
}

\newcommand{\Tref}[1]{Table~\ref{#1}}

\newcommand{\Fref}[1]{Figure~\ref{#1}}
\newcommand{\Sref}[1]{Section~\ref{#1}}

\usepackage[breaklinks=true,bookmarks=false]{hyperref}

\iccvfinalcopy 

\def\iccvPaperID{2980} 

\ificcvfinal\fi

\begin{document}

\title{SPLINE-Net: Sparse Photometric Stereo through \\Lighting Interpolation and Normal Estimation Networks}

\author{\vspace{-22pt}
Qian Zheng$^{1}$\thanks{Authors contributed equally to this work.}\hspace{4pt}\thanks{Corresponding authors.}~~~~~Yiming Jia$^{2\hspace{1pt}\sharp}$\thanks{Work completed while interning at NTU and PKU.}~~~~Boxin Shi$^{3,4\hspace{1pt}*}$~~~~Xudong Jiang$^{1}$~~~Ling-Yu Duan$^{3,4}$~~~Alex C. Kot$^{1}$\vspace{-7pt}}
\affil{\small $^1$School of Electrical and Electronic Engineering, Nanyang Technological University, Singapore \\
$^2$Department of Precision Instrument, Tsinghua University, Beijing, China\\ 
$^3$National Engineering Laboratory for Video Technology, Department of CS, Peking University, Beijing, China \\
$^4$Peng Cheng Laboratory, Shenzhen, China\\
\{zhengqian, exdjiang, eackot\}@ntu.edu.sg, jiaym15@outlook.com, \{shiboxin, lingyu\}@pku.edu.cn
\vspace{-15pt}
}

\maketitle
\ificcvfinal\thispagestyle{empty}\fi

\begin{abstract}
   This paper solves the Sparse Photometric stereo through Lighting Interpolation and Normal Estimation using a generative Network (SPLINE-Net). SPLINE-Net contains a lighting interpolation network to generate dense lighting observations given a sparse set of lights as inputs followed by a normal estimation network to estimate surface normals. Both networks are jointly constrained by the proposed symmetric and asymmetric loss functions to enforce isotropic constrain and perform outlier rejection of global illumination effects. SPLINE-Net is verified to outperform existing methods for photometric stereo of general BRDFs by using only ten images of different lights instead of using nearly one hundred images.
\end{abstract}

\section{Introduction}
\label{sec:intro}

The problem of photometric stereo ~\cite{woodham1980photometric} inversely solves for the radiometric image formation model to recover surface normals from different appearances of objects under various lighting conditions with a fixed camera view. The classic method~\cite{woodham1980photometric} assumes an ideal Lambertian image formation model without global illumination effects (such as inter-reflection and shadows), which deviates from the realistic scenario and prevents photometric stereo from being able to handle real-world objects. To make photometric stereo practical, the major difficulties lie in dealing with objects of general reflectance and global illumination effects. These can be achieved by either exploring analytical Bidirectional Reflectance Distribution Function (BRDF) representations (\emph{e.g.},~\cite{goldman2010shape}) and general BRDF properties (\emph{e.g.},~\cite{shi2014bi}) to model non-Lambertian interactions of lighting and surface normal or suppressing global effects by treating them as outliers (\emph{e.g.},~\cite{wu2010photometric}). Recently, deep learning based approaches are introduced to solve these difficulties by implicitly learning both the image formation process and global illumination effects from training data (\emph{e.g.,}~\cite{chen2018ps,ikehata2018cnn}).

\begin{figure}[t]
{\centering{\includegraphics[width=1\linewidth]{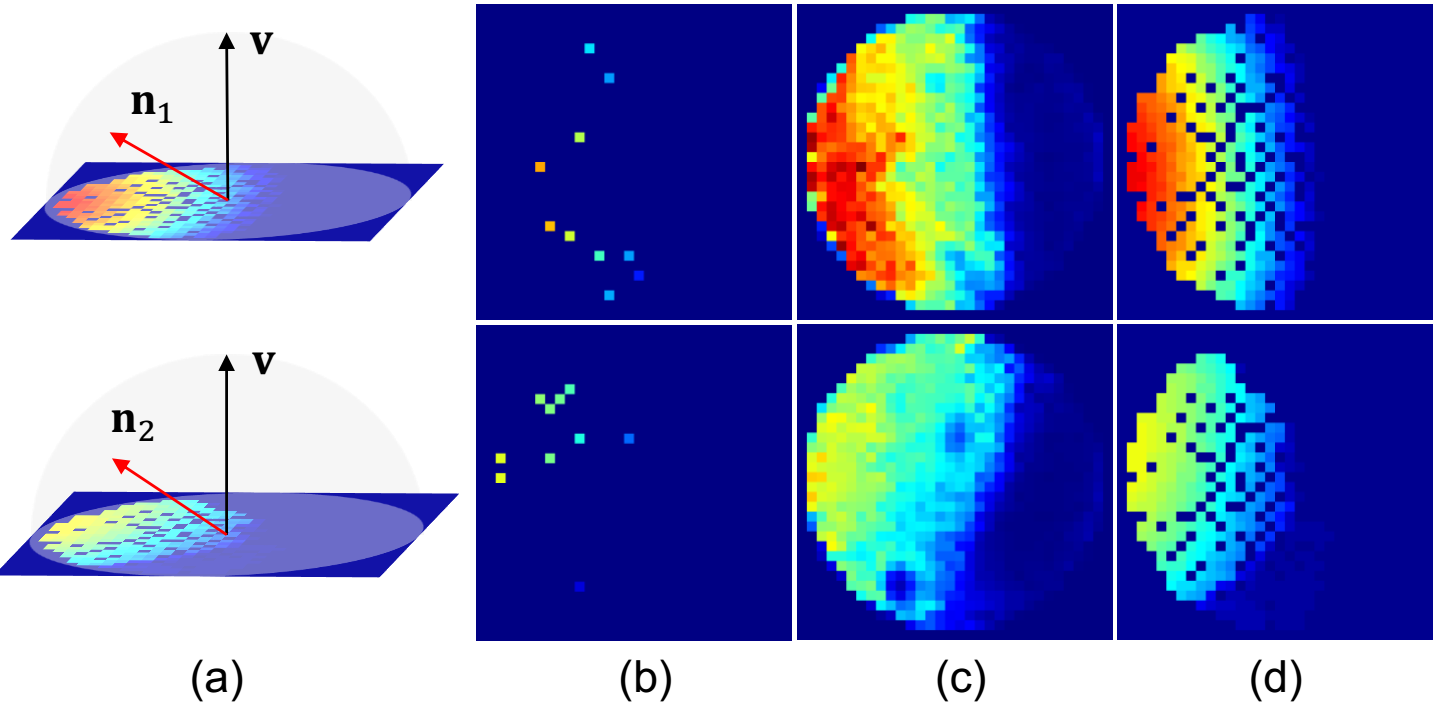}}
\vspace{-15pt}
\caption{An illustration of observation maps corresponding to two surface normals (a brief introduction of observation maps can be found in \Sref{subsec:loss} and~\cite{ikehata2018cnn}).  (a) Two surface normals and their observation maps with dense lights, (b) sparse observation maps with 10 order-agnostic lights, (c) dense observation maps generated by our SPLINE-Net given sparse observation maps in (b) as inputs, and (d) ground truth of dense observation maps with 1000 lights.
We use $\mathbf{v}=(0,0,1)^\top$ to represent viewing direction and $\mathbf{n}$ to represent surface normal in this paper.}
\label{fig:intro}}
\vspace{-10pt}
\end{figure}

According to a comprehensive benchmark evaluation \cite{shi2019benchmark} (including quantitative results for representative methods published before 2016) and the additional results reported in most recent works~\cite{chen2018ps,ikehata2018cnn,zheng2019numerical}, a moderately dense lighting distribution (\emph{e.g.}, around 100 directional lights randomly sampled from the visible hemisphere) is required to achieve reasonably good normal estimation for objects with general materials (\emph{e.g.}, angular error around $10^\circ$ for a shiny plastic).   This is because multi-illumination observations under a dense set of lights are required to fit the parameters in analytic BRDF models~\cite{goldman2010shape}, to analyze general BRDF properties~\cite{shi2014bi}, to observe sufficient inliers and outliers~\cite{wu2010photometric}, and to ensure the convergence of training neural networks~\cite{chen2018ps}.
How to achieve high accuracy in normal estimation for objects given general BRDFs with a sparse set of lights (\emph{e.g.}, 10), which we call sparse photometric stereo in this paper, is still an open yet challenging problem~\cite{shi2019benchmark}. 

In this paper, we propose to solve Sparse Photometric stereo through Lighting Interpolation and Normal Estimation Networks, namely SPLINE-Net. The SPLINE-Net is composed of two sub-networks: the Lighting Interpolation Network (LI-Net) to generate dense observations given a sparse set of input lights and the Normal Estimation Network (NE-Net) to estimate surface normal from the generated dense observations. 
LI-Net takes advantage of a learnable representation for dense lighting called {\it observation map}~\cite{ikehata2018cnn}, and we propose to deal with sparse observation maps as damaged paintings and generate dense observation through inpainting (as shown in~\Fref{fig:intro}\footnote{Note that holes in ground truth of observation maps are produced by the discrete projection from a limited number of lights to a grid with limited resolution (\emph{e.g.}, from 1000 lighting directions to observation maps with size $32\times32$ in this figure).}). NE-Net then follows LI-Net to infer surface normal guided by dense observation maps. To accurately guide the lighting interpolation and normal estimation specially under the photometric stereo context, we propose a {\it symmetric loss} and an {\it asymmetric loss} to explicitly consider general BRDF properties and outlier rejections. More specifically, the symmetric loss is derived according to the property of isotropy for general reflectance, which constrains pixel values on a generated observation map to be symmetrically distributed {w.r.t.} an axis determined by the corresponding surface normal. The asymmetric loss is derived from contaminated observation maps with global illumination effects, which constrains the difference between values of symmetrically distributed pixels to be equal to a non-zero amount. SPINE-Net is validated to achieve superior normal estimation accuracy given a small number of input images (\emph{e.g.}, 10) comparing to state-of-the-art methods using a much larger number (\emph{e.g.}, 96), which greatly relieves data capture and lighting calibration labor for photometric stereo with general BRDFs. The contributions of this paper are two-fold:
\begin{itemize}
   \item We propose the SPLINE-Net to address the problem of photometric stereo with general BRDFs using a small number of images through an integrated learning procedure of lighting interpolation and normal estimation.  
   \item We show how symmetric and asymmetric loss functions can be formulated to facilitate the learning of lighting interpolation and normal estimation with isotropy constraint and outlier rejection of global illumination effects considered.
\end{itemize}

\begin{figure*}[t]
{\centering{\includegraphics[width=1\linewidth]{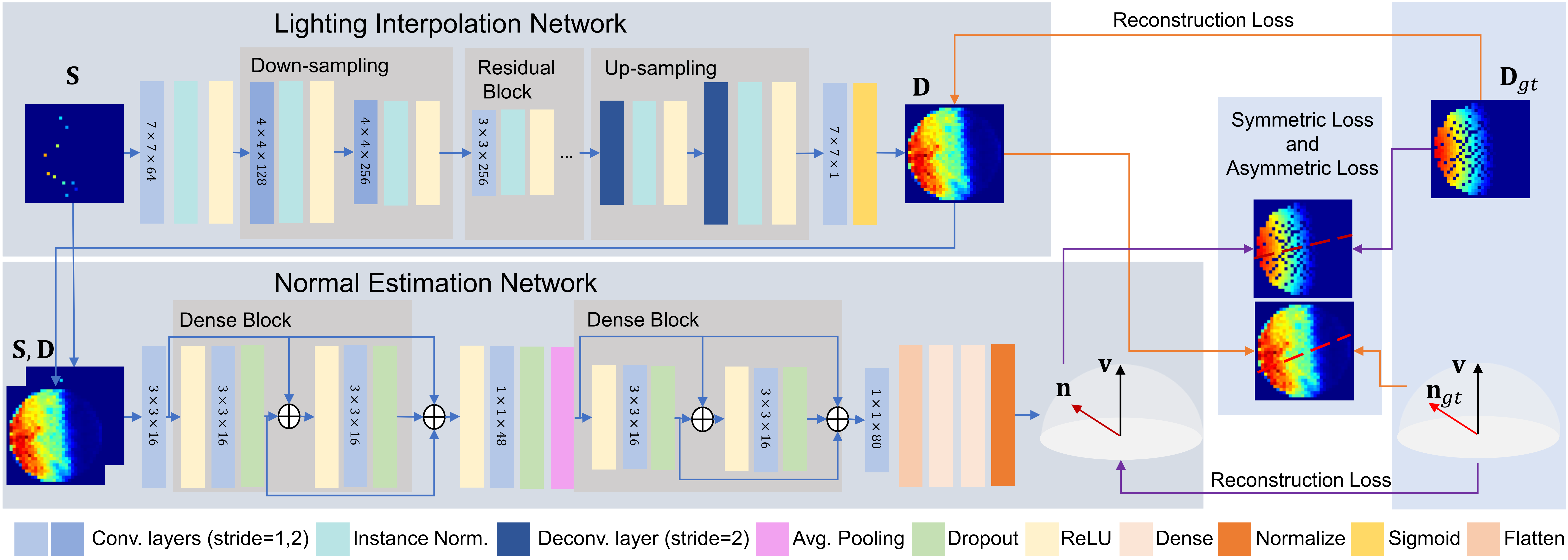}}
\vspace{-12pt}
\caption{The framework of the proposed SPLINE-Net. The lighting interpolation network generates dense observation maps $\mathbf{D}$ given sparse observation maps $\mathbf{S}$ as inputs. The normal estimation network estimates surface normals $\mathbf{n}$ given $\mathbf{S}$ and $\mathbf{D}$ as inputs. Both networks are trained in a supervised manner where ground truth of observation maps $\mathbf{D}_{gt}$ and surface normals $\mathbf{n}_{gt}$ are known.}
\label{fig:framework}}
\vspace{-12pt}
\end{figure*}

\section{Related Works}
\label{sec:related}

In this section, we briefly review traditional methods and deep learning based methods for non-Lambertian photometric stereo with general materials and known lightings.
For other generalizations of photometric stereo, we refer readers to survey papers in~\cite{herbort2011introduction,ackermann2015survey,shi2019benchmark}.

\noindent\textbf{Traditional methods.}
The classical method~\cite{woodham1980photometric} for the problem of photometric stereo is to assume a Lambertian surface reflectance and recover surface normals pixel-wisely. 
Such an assumption is too strong to provide an accurate recovery in real-world due to densely observed non-Lambertian reflectance caused by materials with diverse reflectance and global illumination effects.
In order to address non-Lambertian reflectance from broad classes of materials, modern algorithms attempt to use a mathematically tractable form to describe BRDFs.
Analytic models exploit all available data to fit a nonlinear analytic BRDF, such as Blinn-Phong model~\cite{tozza2016direct}, Torrance-Sparrow model~\cite{georghiades2003incorporating}, Ward model and its variations~\cite{chung2008efficient,goldman2010shape,ackermann2012photometric}, specular spike model~\cite{chen2006mesostructure,yeung2015normal}, and microfacet BRDF model~\cite{chen2017microfacet}. 
Empirical models consider general properties of a BRDF, such as isotropy, monotonicity.
Some basic derivations for isotropy BRDFs are provided in~\cite{alldrin2007toward,tan2011geometry,chandraker2013differential}.
Excellent performance has been achieved by methods based on empirical models, including combining isotropy and monotonicity with visibility constraint~\cite{higo2010consensus},  using isotropic constraint for the estimation of elevation angle~\cite{alldrin2008photometric,shi2012elevation,li2015photometric}, and approximating isotropic BRDFs by bivariate functions~\cite{romeiro2008passive,ikehata2014photometric,shi2014bi}.  
However, most of these methods based on analytic models and empirical models are pixel-wise so that they cannot explicitly consider global illumination effects such as inter-reflection and cast shadows. 
Outlier rejection based methods are developed to suppress global illumination effects by considering them as outliers. 
Earlier works select a subset of Lambertian images from inputs for the accurate recovery of surface normals~\cite{mukaigawa2007analysis,verbiest2008photometric,miyazaki2010median,yu2010photometric,wu2010photometric}.
Recent methods apply robust analysis by assuming non-Lambertian reflectance is sparse~\cite{wu2010robust,ikehata2012robust}.
However, these methods still rely on the existence of a dominant Lambertian reflectance component.

\noindent\textbf{Deep learning based methods.}
Recently, with the great success in both high-level and low-level computer vision tasks achieved by neural networks, researchers have introduced deep learning based methods to solve the problem of photometric stereo.
Instead of explicitly modeling image formation process and global illumination effects as in traditional methods, deep learning based methods attempt to learn such information from data.
DPSN~\cite{santo2017deep} is the first attempt and it uses a deep fully-connect network to regress surface normals from given observations captured under pre-defined lightings in a supervised manner.
However, pre-definition of lightings limits its practicality for photometric stereo where the number of inputs often varies.
PS-FCN~\cite{chen2018ps} is proposed to address such a limitation and handle images under various lightings in an order-agnostic manner by aggregating features of inputs using the max-pooling operation.
CNN-PS~\cite{ikehata2018cnn} is another work to accept order-agnostic inputs by introducing observation map, which is a fixed shape representation invariant to inputs. 
Besides neural networks trained in a supervised manner, Taniai and Maehara~\cite{taniai2018neural} presented an unsupervised learning framework where surface normals and BRDFs are recovered by minimizing a reconstruction loss between inputs and images synthesized based on a rendering equation.

Only a few earlier works address the problem of photometric stereo with general reflectance using a small number of images in the literature (\emph{e.g.}, analytic model based method~\cite{goldman2010shape}, shadow analysis based method~\cite{chandraker2007shadowcuts}).
Our paper revisits this problem due to its low costs for the labor of data capture and lighting calibration.

\section{The Proposed SPLINE-Net}
\label{sec:rrnet}

In this section, we introduce our solution to the problem of photometric stereo with general reflectance using a small number of images.
We first present the framework of our SPLINE-Net in \Sref{subsec:framework}.
Then we detail the symmetric loss and the asymmetric loss in \Sref{subsec:loss}.

\subsection{Framework}
\label{subsec:framework}

As illustrated in~\Fref{fig:framework}, our SPLINE-Net, which consists of a Lighting Interpolation Network (LI-Net) and a Normal Estimation Network (NE-Net), is optimized in a supervised manner.
LI-Net (represented as a regression function $f$) interpolates dense observation maps $\mathbf{D}$ from sparse observation maps $\mathbf{S}$ (\emph{i.e.}, sparse sets of lights), 
\begin{equation}
f:\mathbf{S}\rightarrow\mathbf{D}.
\end{equation}
Such densely interpolated observation maps $\mathbf{D}$ are then concatenated to original inputs $\mathbf{S}$ and help estimate surface normals $\mathbf{n}$ in NE-Net (represented as a regression function $g$)
\begin{equation}
g:\mathbf{S}, \mathbf{D}\rightarrow\mathbf{n}.
\end{equation}

LI-Net and NE-Net are trained in an alternating iteratively manner, where fixing one network when optimizing the other.
Specifically, we update LI-Net once after updating NE-Net five times.
The loss function for each network composes of a reconstruction loss, a symmetric loss, and an asymmetric loss.

\noindent\textbf{Lighting Interpolation Network.}
Inspired by a recent work~\cite{chen2019self} that suggests inferring lighting information as intermediate results can benefit the estimation of the surface normal, we design the LI-Net to generate observation maps regarding dense lighting directions.
The basic idea is to inpaint sparse observation maps and obtain dense ones, based on learnable properties~(\emph{e.g.}, spatial continuity).
LI-Net is designed using an encoder-decoder structure due to its excellent image generation capacity~\cite{johnson2016perceptual, zhu2017unpaired}.
The loss function of LI-Net $\mathcal{L}_f$ is formulated as,
\begin{equation}
\label{equ:f}
\mathcal{L}_f =\mathcal{L}^{rec}_f+\lambda_{s}\mathcal{L}^{s}_f+\lambda_{a}\mathcal{L}^{a}_f.
\end{equation}
The reconstruction loss $\mathcal{L}^{rec}_f$ is defined as\footnote{We experimentally find L1 and L2 distances provide similar results and here we compute reconstruction loss $\mathcal{L}^{rec}_f$ using L1 distance.},
{\small
\begin{equation}
\label{equ:ff}
\mathcal{L}^{rec}_f = \arccos(\mathbf{n}^\top\mathbf{n}_{gt}) + |\mathbf{D}-\mathbf{D}_{gt}|_1 + |\mathbf{M}_s\circ(\mathbf{D}-\mathbf{D}_{gt})|_1 ,
\end{equation}}
\hspace{-4pt}where $\mathbf{D} = f(\mathbf{S}), \mathbf{n}=g(\mathbf{S},\mathbf{D})$, $\mathbf{n}_{gt}$ and $\mathbf{D}_{gt}$ are ground truth of a surface normal and its corresponding dense observation map, respectively, $\mathbf{M}_s$ is a binary mask indicating positions of non-zero value of $\mathbf{S}$, $\circ$ represents element-wise multiplication.
$\mathcal{L}^{s}_f$ and $\mathcal{L}^{a}_f$ are our symmetric and asymmetric loss to be introduced in \Sref{subsec:loss}.

\noindent\textbf{Normal Estimation Net.}
We use the same architecture as that in~\cite{ikehata2018cnn} (a variation of DenseNet~\cite{huang2017densely}) for NE-Net due to its excellent capacity to model the relation between observation maps and surface normals.
The loss function of NE-Net is formulated as,
\begin{equation}
\label{equ:g}
\begin{split}
&\mathcal{L}_g = \mathcal{L}^{rec}_g +\lambda_{s}\mathcal{L}^{s}_g+\lambda_{a}\mathcal{L}^{a}_g,\\
\end{split}
\end{equation}
where $\mathcal{L}^{s}_g$ and $\mathcal{L}^{a}_g$ are symmetric loss and asymmetric loss, and reconstruction loss is,
\begin{equation}
\label{equ:gg}
\begin{split}
&\mathcal{L}^{rec}_g = \arccos(\mathbf{n}^\top\mathbf{n}_{gt}).\\
\end{split}
\end{equation}

\begin{figure}[t]
{\centering{\includegraphics[width=1\linewidth]{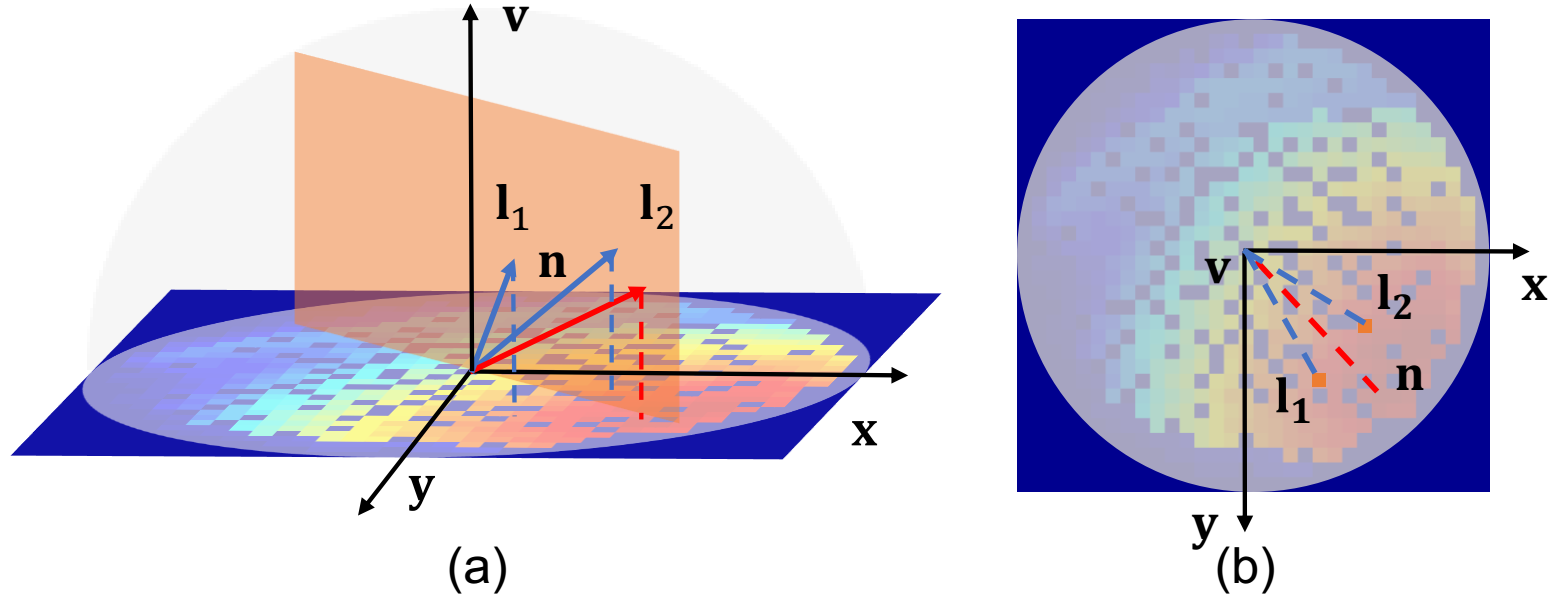}}
\vspace{-15pt}
\caption{An illustration of an orthogonal projection from a hemisphere surface to its base (gray) and an interpretation of isotropy for a dense observation map. (a) Front view: $\mathbf{l}_1$ and $\mathbf{l}_2$ represent two lighting directions which are symmetric w.r.t. the plane spanned by viewing direction $\mathbf{v}$ and surface normal $\mathbf{n}$~(orange plane). (b) Top view: irradiance values (orange dots), which are projected by $\mathbf{l}_1$ and $\mathbf{l}_2$, are numerically equal due to isotropy.}
\label{fig:proj}}
\vspace{-10pt}
\end{figure}

\subsection{Symmetric Loss and Asymmetric Loss}
\label{subsec:loss}

In this section, we first revisit the observation map from~\cite{ikehata2018cnn}.
Then, we further investigate its characteristics by considering isotropy and global illumination effects.
Finally, we present symmetric and asymmetric loss functions.

\noindent\textbf{Observation maps.}
As introduced in~\cite{ikehata2018cnn}, each point on a surface normal map corresponds to an observation map (as shown in \Fref{fig:proj}~(a)).
Elements on such a map describe observed irradiance values under different lighting directions.
These lighting directions are mapped to positions of elements, which is an orthogonal projection.
As illustrated in \Fref{fig:proj}~(a), a dense observation map can be regarded as generated by projecting a hemisphere surface to its base plane, where each point on the hemisphere surface represents a direction of lighting and its projecting value describes an observed irradiance value under such a light.
Such a projecting relation motivates us to introduce isotropy to narrow the solution space of our SPLINE-Net.

\begin{figure}[t]
{\centering{\includegraphics[width=1\linewidth]{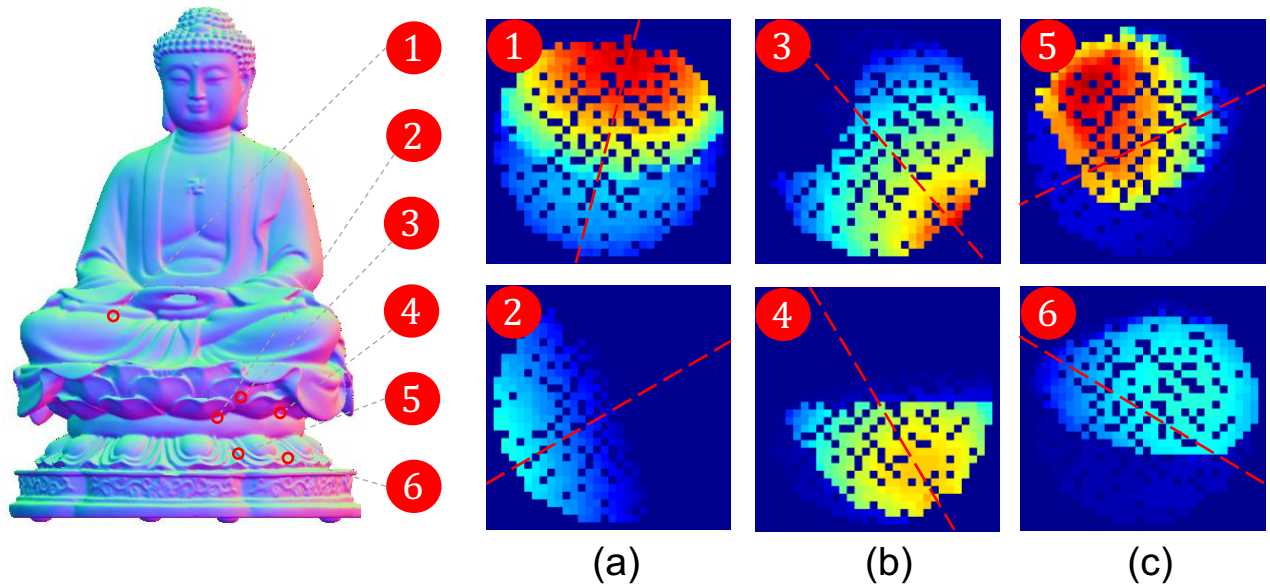}}
\vspace{-15pt}
\caption{Six dense observation maps of data {\sc buddha} from {\sc CyclesPS}~\cite{ikehata2018cnn} with (a) isotropic reflectance, (b) cast shadows, and (c) inter-reflection. Red-dotted lines indicate directions of their corresponding surface normals.}
\label{fig:pattern}}
\vspace{-10pt}
\end{figure}

\noindent\textbf{Isotropic BRDFs and global illumination effects.}
Isotropy BRDFs $\rho(\mathbf{n}^\top\boldsymbol{\ell},\mathbf{n}^\top\mathbf{v},\mathbf{v}^\top\boldsymbol{\ell})$ for general materials have the property that reflectance values are numerically equal if lighting directions are symmetric about the plane spanned by $\mathbf{v}$ and $\mathbf{n}$ as shown in \Fref{fig:proj}~(a).
Considering the relation of the one-to-one mapping between lighting directions and positions of observed irradiance values, these values are numerically equal if their positions are symmetrically distributed regarding to the axis projected by surface normals in observation maps, as shown by \Fref{fig:proj}~(b).
However, such symmetric patterns can be destroyed by global illumination effects because observation maps are pixel-wisely generated.
Therefore, unpredictable shapes produce cast shadows or inter-reflection which can lead to sudden changes of irradiance values on observation maps.
\Fref{fig:pattern} illustrates examples of isotropy, cast shadow, inter-reflection.

\begin{table*}[ht]
\begin{center}
\caption{Quantitative comparisons in terms of angular error on {\sc CyclesPS-Test} dataset~\cite{ikehata2018cnn}. Results of three shapes ({\sc paperbowl}, {\sc sphere}, {\sc turtle}) with metallic ({\sc M}) and specular ({\sc S}) materials are reported. 
Note that all results are averaged over 100 random trials. 
For subset {\sc L17} (left) and subset {\sc L305} (right), results over 6 different data are averaged (Avg.) for each method.
}
\label{tab:all_syn}
{\footnotesize
\begin{tabular}{|c|c c| c c |c c| c ||c c| c c| c c| c| }
\hline
& \multicolumn{2}{c|}{\sc Paperbowl}&\multicolumn{2}{c|}{\sc Sphere} & \multicolumn{2}{c|}{\sc Turtle} & \multirow{2}{*}{Avg.} & \multicolumn{2}{c|}{\sc Paperbowl}&\multicolumn{2}{c|}{\sc Sphere} & \multicolumn{2}{c|}{\sc Turtle} & \multirow{2}{*}{Avg.} \\
\cline{2-7}
\cline{9-14}
 &{\sc M}&{\sc S}&{\sc M}&{\sc S}&{\sc M}&{\sc S}&&{\sc M}&{\sc S}&{\sc M}&{\sc S}&{\sc M}&{\sc S}&\\
\hline
LS~\cite{woodham1980photometric} &{\cca{41.47}}    &{\cca{35.09}}&    {\cca{18.85}}&    {\cca{10.76}}    &{\cca{27.74}}    &{\cca{19.89}}    &{\cca{25.63}}&    {\cca{43.09}}&    {\cca{37.36}}&    {\cca{20.19}}    &{\cca{12.79}}    &{\cca{28.51}}&    {\cca{21.76}}&    {\cca{27.28}}\\
IW12~\cite{ikehata2012robust}& {\cca{46.68}} &    {\cca{33.86}} &    {\cca{16.77}}     &{\bf\cca{2.23}} &    {\cca{31.83}} &    {\cca{12.65}}     &{\cca{24.00}}     &{\cca{48.01}}     &{\cca{37.10}}&     {\cca{21.93}} &    {\bf\cca{3.19}} &    {\cca{34.91}}    &{\cca{16.32}} &    {\cca{26.91}} \\
ST14~\cite{shi2014bi} &{\cca{42.94}}    &{\cca{35.13}}    &{\cca{22.58}}    &{\cca{4.18}}    &{\cca{34.30}}    &{\cca{17.01}}&    {\cca{26.02}}&    {\cca{44.44}}    &{\cca{37.35}}&    {\cca{25.41}}    &{\cca{4.89}}    &{\cca{36.01}}&    {\cca{19.06}}    &{\cca{27.86}}\\
IA14~\cite{ikehata2014photometric}&{\cca{48.25}} &    {\cca{43.51}}     &{\cca{18.62}}&     {\cca{11.71}} &    {\cca{30.59}}     &{\cca{23.55}} &{\cca{29.37}}&     {\cca{49.01}}     &{\cca{45.37}} &    {\cca{21.52}} &    {\cca{13.63}} &    {\cca{32.82}}     &{\cca{26.27}}&    {\cca{31.44}} \\
CNN-PS~\cite{ikehata2018cnn}& {\cca{37.14}}& {\cca{23.40}}&    {\cca{17.44}}&    {\cca{6.99}}    &{\cca{22.86}}&    {\cca{10.74}}&    {\cca{19.76}}&    {\cca{38.45}}&    {\cca{26.90}}&    {\cca{18.25}}    &{\cca{9.04}}&    {\cca{23.91}}&    {\cca{14.36}}&    {\cca{21.82}}\\
SPLINE-Net  & {\bf\cca{29.87}}    &{\bf\cca{18.65}}    &{\bf\cca{6.59}}    &{\cca{3.82}}    &{\bf\cca{15.07}}&    {\bf\cca{7.85}}&    {\bf\cca{13.64}}    &{\bf\cca{33.99}}    &{\bf\cca{23.15}}    &{\bf\cca{9.21}}&    {\cca{6.69}}&    {\bf\cca{17.35}}&    {\bf\cca{12.01}}&    {\bf\cca{17.07}}\\                                                           
\hline
\end{tabular}}
\end{center}
\vspace{-20pt}
\end{table*}

\noindent\textbf{Symmetric and asymmetric loss functions.}
In order to further narrow solution spaces for the lighting interpolation of dense observation maps to facilitate accurate estimation of the surface normal, we propose symmetric and asymmetric loss functions to exploit above observations for LI-Net and NE-Net.
More specifically, given a dense observation map $\mathbf{D}$ and its corresponding surface normal $\mathbf{n}$, the symmetric loss forces the isotropic properties of general BRDFs which are valid on various real-world reflectance. 
That is, it constrains irradiance values, which are symmetrically distributed w.r.t. an axis determined by its surface normal (red-dotted lines in \Fref{fig:pattern}), to be numerically equal,
\begin{equation}
\label{equ:symetric}
\begin{split}
&\mathcal{L}_s = \mathcal{L}_s(\mathbf{D},\mathbf{n}) = |\mathbf{D}- r(\mathbf{D}, \mathbf{n})|_1,
\end{split}
\end{equation}
where function $r(\mathbf{D}, \mathbf{n})$ mirrors the observation map $\mathbf{D}$ {w.r.t.} the axis determined by $\mathbf{n}$.
Different from symmetric loss, the asymmetric loss models the asymmetric pattern brought by outliers such as global illuminations.
It constrains the difference between values of symmetrically distributed pixels to be equal to a non-zero amount $\eta$,
\begin{equation}
\label{equ:asymetric}
\begin{split}
\mathcal{L}_a = \mathcal{L}_a(\mathbf{D},\mathbf{n}) &=\big||\mathbf{D} - r(\mathbf{D}), \mathbf{n}|_1-\eta\big|_1 \\
 &+  \lambda_c \big||p(\mathbf{D}) - r(p(\mathbf{D}), \mathbf{n})|_1-\eta\big|_1,
\end{split}
\end{equation}
where $\lambda_c$ is a weight parameter, function $p(\cdot)$ performs an average pooling operation  with stride of 2 to ensure spatial continuity of observation maps.
Empirically, we set $\eta=1, \lambda_c=50$ for all experiments.
Both of $\mathcal{L}_s$ and $\mathcal{L}_a$ aim to better fit observations of symmetric and asymmetric patterns (as illustrated in \Fref{fig:pattern}) during training.
We integrate symmetric and the asymmetric loss functions to optimize LI-Net by setting,
\begin{equation}
\label{equ:saf}
\begin{split}
&\mathcal{L}^{s}_f = \mathcal{L}_s(\mathbf{D},\mathbf{n}_{gt}), \\
&\mathcal{L}^{a}_f = \mathcal{L}_a(\mathbf{D},\mathbf{n}_{gt}),
\end{split}
\end{equation}
and to optimize NE-Net by setting,
\begin{equation}
\label{equ:sag}
\begin{split}
&\mathcal{L}^{s}_g = \mathcal{L}_s(\mathbf{D}_{gt},\mathbf{n}), \\
&\mathcal{L}^{a}_g = \mathcal{L}_a(\mathbf{D}_{gt},\mathbf{n}).
\end{split}
\end{equation}
We empirically set $\lambda_s=2\times10^{-2}, \lambda_a=2\times10^{-5}$ due to the fact that global illumination effects are always observed for small regions of real-world objects.

\section{Experiments}

\noindent\textbf{Settings and implementation details.}
A recent survey work~\cite{shi2019benchmark} implies that photometric stereo with general BRDFs shows significant performance drop if only around 10 images are provided. 
Therefore, we define 10 as the number of sparse lights and use 10 randomly sampled lights as inputs for both training and testing. 
The resolution of our observation map is set to $32\times32$ and the batch size is set to $128$, which are the same as those in~\cite{ikehata2018cnn}, for easier comparisons.
Adam optimizer is used to optimize our networks with parameters~$\beta_1=0.9$ and $\beta_2=0.999$. 

\noindent\textbf{Datasets and evaluation.}
We use {\sc CyclesPS} data provided by~\cite{ikehata2018cnn} as our training data.
There are 45 training data including 15 shapes with 3 categories of reflectance~(diffuse, metallic, and specular).
Our testing sets are built based on public evaluation datasets.
To cover as many lighting conditions as possible, we construct 100 instances for each testing data from these datasets and each instance contains images illuminated under 10 randomly selected lights.
Quantitative results are averaged of mean angular errors and metric used in the paper is the angular error in degrees (unit is omitted in all tables).

\noindent\textbf{Compared methods.}
We compare with five methods, including the least squares based Lambertian photometric stereo method LS~\cite{woodham1980photometric}, an outlier rejection based method IW12~\cite{ikehata2012robust}, two state-of-the-art methods exploring general BRDF properties ST14~\cite{shi2014bi} and IA14~\cite{ikehata2014photometric}, and a deep learning based method CNN-PS~\cite{ikehata2018cnn}.
We re-train CNN-PS~\cite{ikehata2018cnn} by taking 10 observed irradiance values as inputs.\footnote{Considering the overall quantitative results of CNN-PS~\cite{ikehata2018cnn} with default settings (taking 50 observed irradiance values as inputs) on {\sc CyclesPS-Test}~\cite{ikehata2018cnn} ({\sc L17}: 31.08$^\circ$ and {\sc L305}: 34.90$^\circ$) and {\sc DiLiGenT}~\cite{shi2019benchmark} (14.10$^\circ$), we report results re-trained by our setting. PS-FCN~\cite{chen2018ps} is not compared due to different requirements of training data.}

\begin{figure*}[h]
{\centering{\includegraphics[width=1\linewidth]{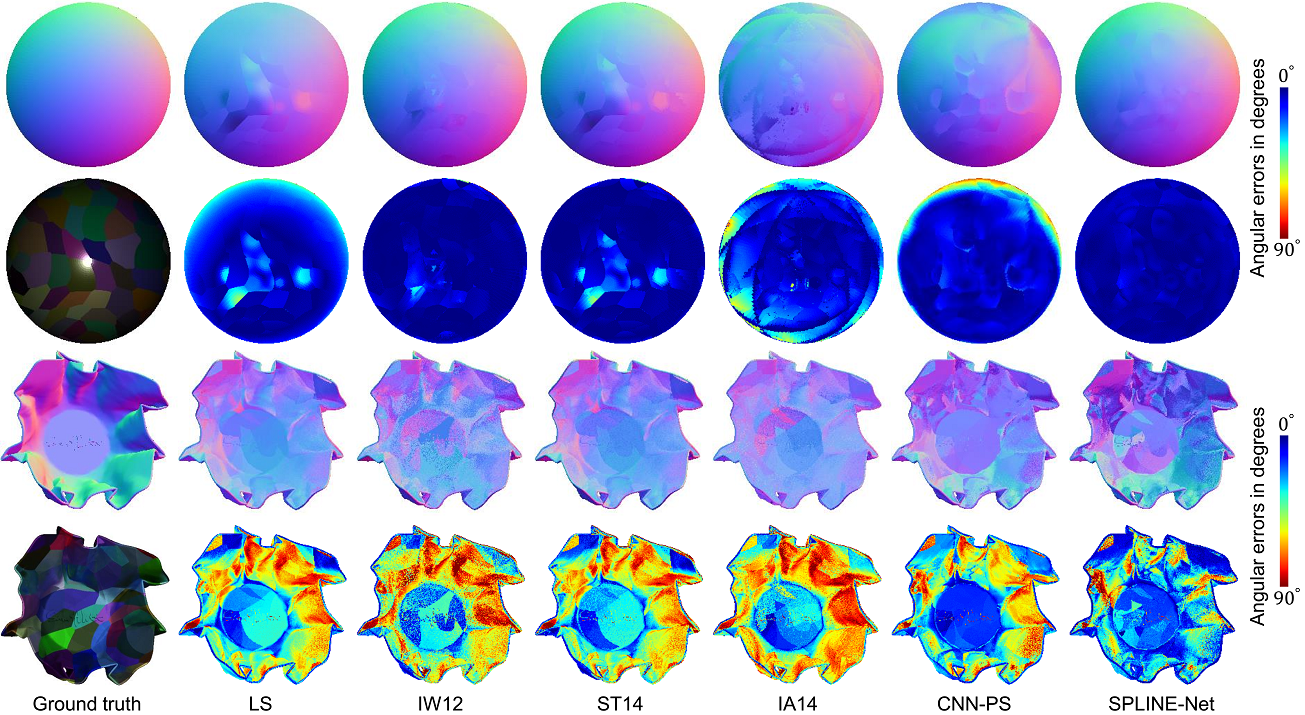}}
\vspace{-10pt}
\caption{Comparisons of normal maps and angular error maps (in degrees) on {\sc sphere} with specular materials (top) and {\sc paperbowl} with metallic materials (bottom) from {\sc L17}, {\sc CyclesPS-Test}~\cite{ikehata2018cnn}.}
\label{fig:visual_syn}}
\vspace{-0pt}
\end{figure*}

\begin{table*}[t]
\begin{center}
\caption{Quantitative comparisons in terms of angular error on {\sc DiLiGenT} dataset~\cite{shi2019benchmark}. Note that all results are averaged over 100 random trials. Results over 10 different data are averaged (Avg.) for each method. 
}
\label{tab:all_real}
{\footnotesize
\begin{tabular}{|c|c c c c c c c c c c|c |}
\hline
Methods &    {\sc Ball}     &{\sc Bear}    &{\sc Buddha}&    {\sc Cat}    &{\sc Cow}     &{\sc Goblet}    &{\sc Harvest}    &{\sc Pot1}    &{\sc Pot2}    &{\sc Reading}    & Avg.\\
\hline
LS~\cite{woodham1980photometric} &{\ccb{4.41}}     &{\ccb{9.05}} &    {\ccb{15.62}} &    {\ccb{9.03}}     &{\ccb{26.42}} &    {\ccb{19.59}} &    {\ccb{31.31}} &    {\ccb{9.46}} &    {\ccb{15.37}} &    {\ccb{20.16}} &    {\ccb{16.04}}  \\
IW12~\cite{ikehata2012robust}&{\bf\ccb{3.33}} &    {\ccb{7.62}} &    {\ccb{13.36}} &    {\ccb{8.13}} &    {\ccb{25.01}} &{\ccb{18.01}} &    {\ccb{29.37}} &    {\ccb{8.73}} &    {\ccb{14.60}}     &{\ccb{16.63}}     &{\ccb{14.48}} \\
ST14~\cite{shi2014bi} &{\ccb{5.24}}     &{\ccb{9.39}}     &{\ccb{15.79}} &    {\ccb{9.34}}&     {\ccb{26.08}}     &{\ccb{19.71}}     &{\ccb{30.85}} &    {\ccb{9.76}} &    {\ccb{15.57}}     &{\ccb{20.08}}     &{\ccb{16.18}} \\
IA14~\cite{ikehata2014photometric}&{\ccb{12.94}} &    {\ccb{16.40}}     &{\ccb{20.63}} &   {\ccb{15.53}} &    {\ccb{18.08}}     &{\ccb{18.73}}&     {\ccb{32.50}} &   {\bf\ccb{6.28}}     &{\ccb{14.31}}     &{\ccb{24.99}}&     {\ccb{19.04}} \\
CNN-PS~\cite{ikehata2018cnn}&{\ccb{17.86}} &    {\ccb{13.08}}&     {\ccb{19.25}}     &{\ccb{15.67}} &    {\ccb{19.28}} &    {\ccb{21.56}} &    {\ccb{21.52}}     &{\ccb{16.95}} &    {\ccb{18.52}}&    {\ccb{21.30}} &    {\ccb{18.50}} \\
\hline
Nets \emph{w/o} loss &{\ccb{6.06}} &   {\ccb{7.01}} &   {\ccb{10.69}} &{\ccb{8.38}}     &{\ccb{10.39}} &    {\ccb{11.37}}     &{\bf\ccb{19.02}}     & {\ccb{9.42}}&    {\ccb{12.34}} &    {\ccb{16.18}} &{\ccb{11.09}} \\
Nets with $\mathcal{L}^s$ &    {\ccb{5.04}} &    {\bf\ccb{5.89}} &    {\ccb{10.11}} &    {\ccb{7.79}} &    {\ccb{9.38}} &    {\ccb{10.84}} &    {\ccb{19.03}} &    {\ccb{8.91}}&    {\bf\ccb{11.47}}     &{\bf\ccb{15.87}} &    {\ccb{10.43}} \\
SPLINE-Net&{\ccb{4.96}} &    {\ccb{5.99}}&     {\bf\ccb{10.07}} &    {\bf\ccb{7.52}} &    {\bf\ccb{8.80}} &    {\bf\ccb{10.43}} &    {\ccb{19.05}} &    {\ccb{8.77}}&     {\ccb{11.79}}     &{\ccb{16.13}}&    {\bf\ccb{10.35}} \\
\hline
\end{tabular}}
\end{center}
\vspace{-20pt}
\end{table*}

\subsection{Synthetic Data}
\label{subsec:syn}

{\sc CyclesPS-Test} is a testing dataset from~\cite{ikehata2018cnn}, which consists of two subsets (denoted as `{\sc L17}' and `{\sc L305}').
Each of subsets contains three shapes and these shapes are rendered by reflectance of `metallic' and `specular'.\footnote{`Metallic' and `specular' reflectance are generated by controlling 11 parameters from Disney's principled BSDF~\cite{burley2012physically}, which are detailed in~\cite{ikehata2018cnn}.}
For all the 12 data, we construct 100 instances (each with 10 randomly selected images) to build our testing set.

As can be observed from~\Tref{tab:all_syn}, metallic materials are more challenging as compared with specular materials.
Even for the simple shape ({\sc sphere}) containing few global illumination effects, all methods fail to estimate accurate surface normals for metallic materials.
The performance advantage of our method is superior, \emph{i.e.}, the overall performance is much better than the second best one.
Interestingly, two traditional methods, IW12~\cite{ikehata2012robust} and ST14~\cite{shi2014bi}, outperform other methods on data {\sc sphere} with specular materials.
However, their performance degrades for complex shapes ({\sc paperbowl} and {\sc turtle}), while our method consistently achieves the best performance.

Visual comparisons in \Fref{fig:visual_syn} further validates the effectiveness of our method.
It deals with specular reflectance more robustly ({\sc sphere}) and consistently produces the best estimation for most regions on a more complex shape ({\sc paperbowl}).
The superior performance shows the effectiveness of our method to address the problem of photometric stereo with general BRDFs using a small number of images.

\subsection{Real Data}
\label{subsec:real}
The benchmark dataset {\sc DiLiGenT}~\cite{shi2019benchmark} consists of 10 real data. 
Similarly, for each of these 10 data, we construct 100 instances, each of which contains 10 randomly selected images, to build the testing set.

As can be found from quantitative results in \Tref{tab:all_real}, our method demonstrates obvious superiority for most data except for {\sc ball} and {\sc pot1}, which get similar or worse results as compared to two traditional methods LS~\cite{woodham1980photometric} and IW12~\cite{ikehata2012robust}.
The reason is these two data are diffuse-dominant so that traditional methods with Lambertian assumption fit well even for a small number of observed irradiance values.
However, our data-driven approach considers general reflectance and global illumination effects evenly during model optimization and hence may underfit Lambertian surfaces with simple shapes.
Unlike excellent performance on synthetic data, CNN-PS~\cite{ikehata2018cnn} achieve less accuracy on real data.
We think that the reason is mainly due to the problem of overfitting during training, \emph{i.e.}, synthetic data for testing are constructed in a similar manner as training data.
Our method achieves the best accuracy for most of data, such as {\sc cow} (metallic paint materials) and {\sc goblet}, {\sc harvest} (most of regions contain inter-reflection or cast shadows).

\begin{figure*}[t]
{\centering{\includegraphics[width=1\linewidth]{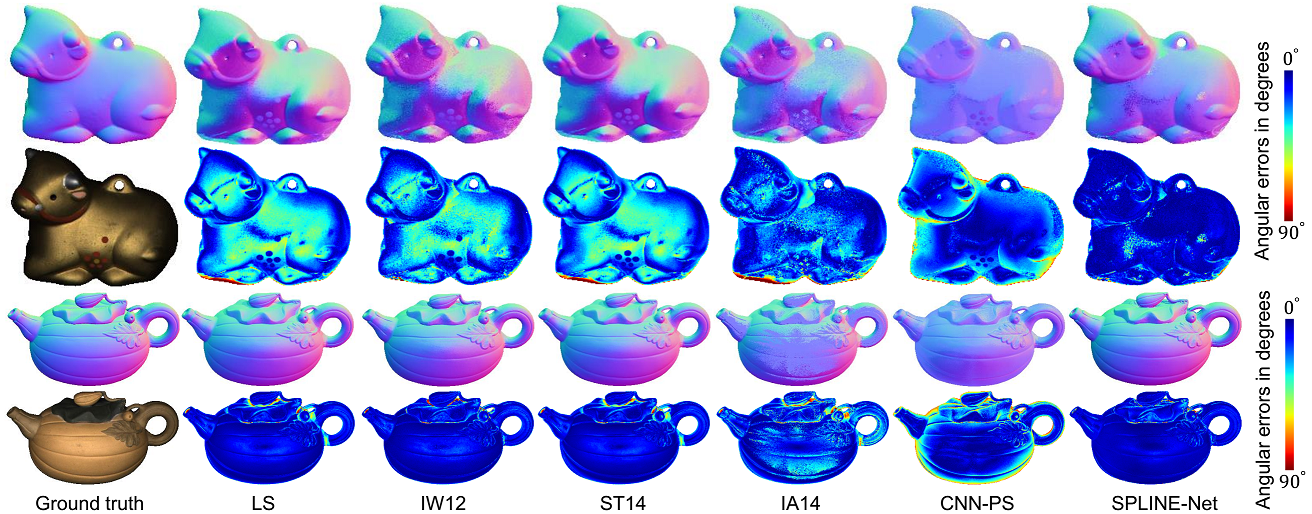}}
\vspace{-15pt}
\caption{Comparisons of normal maps and angular error maps (in degrees) on {\sc cow} (top) and {\sc pot1} (bottom) from {\sc DiLiGenT}~\cite{shi2019benchmark}. }
\label{fig:visual_real}}
\vspace{-15pt}
\end{figure*}

Visual comparisons on data {\sc cow} and {\sc pot1} are shown in \Fref{fig:visual_real}.
Our method provides much more accurate results for metallic materials ({\sc cow}) which is consistent with the results of synthetic data (\Tref{tab:all_syn}).
Most of the compared methods and our method achieve accurate estimation for center regions of {\sc pot1}, however, our method gets a significant advantage for boundaries (\emph{e.g.}, regions of spout and kettle-holder) containing inter-reflection or cast shadows.

\subsection{Robustness against Noise}

\noindent\textbf{Lights.} It is inevitable to introduce errors during lighting calibration in practice.
To show the robustness of our method against errors in light source directions, we manually add noise to lights in {\sc DiLiGenT}~\cite{shi2019benchmark} and test the quantitative performance regarding different levels of noise.
As can be observed in \Tref{tab:noise}, our method is able to produce accurate results when the noise is small ($\sigma=2$). 
However, its performance drops with higher levels of noise.
The robustness of our method to small noise is mainly due to our training data, \emph{i.e.}, small random noise is added to each light when creating training dataset as introduced in~\cite{ikehata2018cnn}.

\begin{table}[t]
\begin{center}
\caption{The performance of SPLINE-Net on {\sc DiLiGenT}~\cite{shi2019benchmark} against different standard deviations ($\sigma$) of Gaussian noise ($\mu=0$) added to lighting directions of inputs. }
\label{tab:noise}
{\footnotesize
\begin{tabular}{|c|c |c| c| c |c|}
\hline
$\sigma$ (in degrees)        & 0 (no noise)& 2& 4    & 6    & 8\\
\hline
Performance    &10.35&11.21&12.14&13.76&15.94\\
\hline
\end{tabular}}
\vspace{-26pt}
\end{center}
\end{table}

\begin{figure}[t]
{
{\centering{\includegraphics[width=1\linewidth]{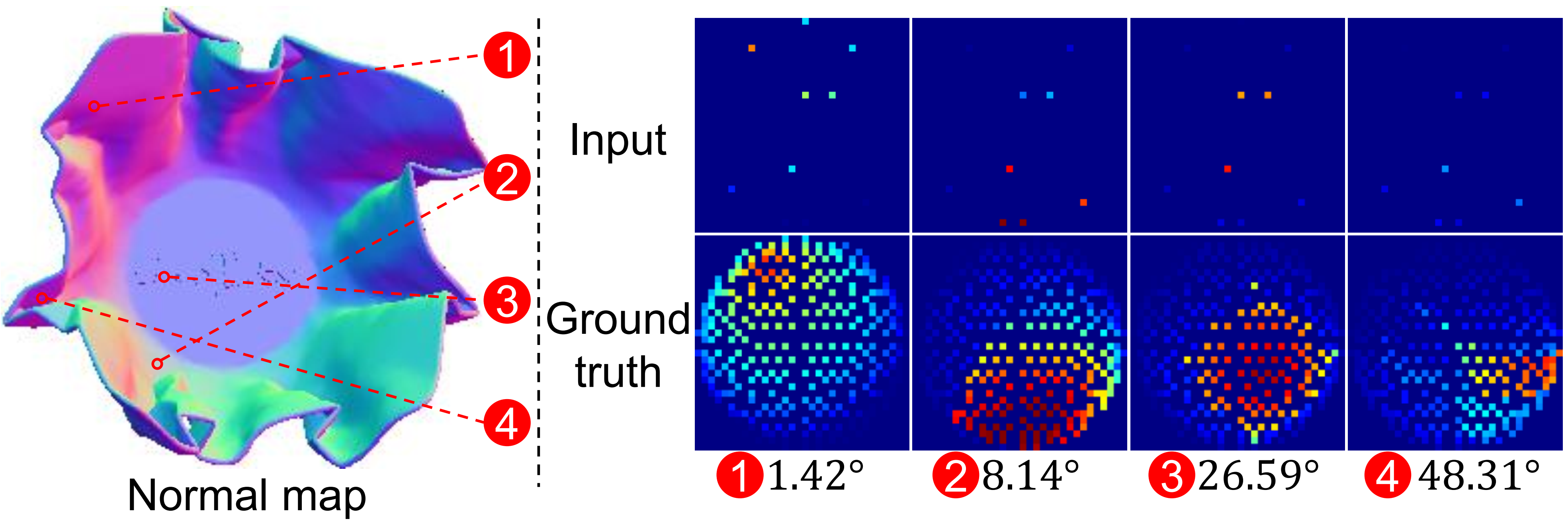}}
\vspace{-15pt}
\caption{Left: a normal map and its four points. Right: four pairs of observation maps (inputs and their ground truth) corresponding to points on the left;
numbers below are errors of estimated surface normal given corresponding inputs. Zoom in for a better view.}
\label{fig:noise}}
\vspace{-10pt}
}
\end{figure}

\noindent\textbf{Inputs.} Because of the sparsity of inputs, the performance of our method is affected by inputs due to outliers resulting from inter-reflection and shadows. 
\Fref{fig:noise} (right) illustrates this consideration.
As can be found, when the number of outliers of inputs increases (the number of dark dots in inputs increases), the error becomes larger accordingly. 

\subsection{Ablation Studies}
\label{subsec:ablation}

In this section, we perform ablation studies to further investigate the contribution of important components in SPLINE-Net.
Considering the same network structure of NE-Net and that in CNN-PS~\cite{ikehata2018cnn} and the fact that our SPLINE-Net is composed of LI-Net and NE-Net,
we compare our SPLINE-Net without symmetric loss or asymmetric loss (denoted as `Nets \emph{w/o} loss') with CNN-PS~\cite{ikehata2018cnn} to validate the effectiveness of LI-Net. 
By comparing the performance of Nets \emph{w/o} loss with that with an additional symmetric loss (denoted as `Nets with $\mathcal{L}^s$'), we validate the effectiveness of enforcing isotropy property.
The comparison performed between Nets with $\mathcal{L}^s$ and SPLINE-Net is to verify the effectiveness of the consideration of global illumination effects.
The same settings and the same testing set as those in \Sref{subsec:real} is used for evaluation in this section.

\begin{figure}[t]
{\centering{\includegraphics[width=1\linewidth]{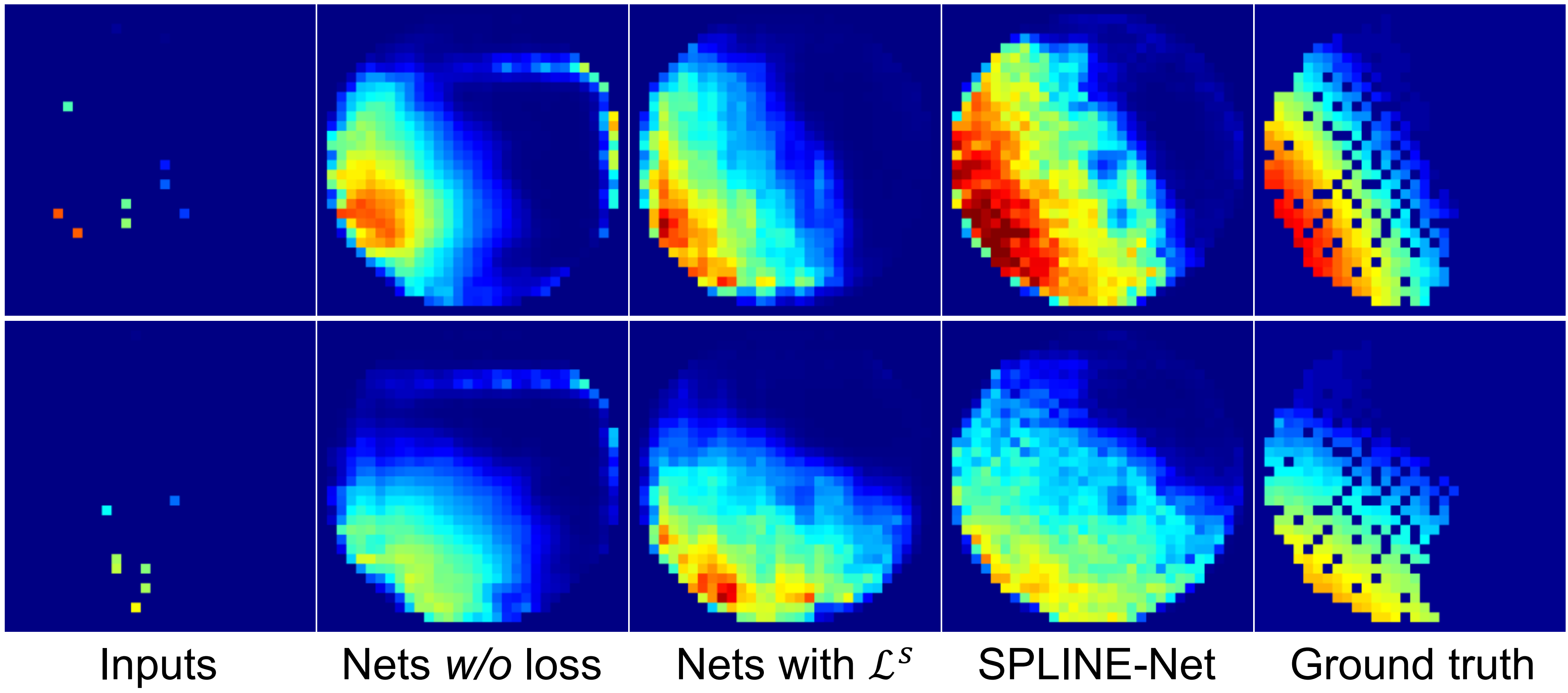}}
\vspace{-15pt}
\caption{An illustration of observation maps. From left to right columns: inputs (10 order-agnostic lights), maps generated by SPLINE-Net {\it w/o} our loss, by SPLINE-Net with symmetric loss $\mathcal{L}^s$, by SPLINE-Net, and ground truth (1000 lights).}
\label{fig:maps}}
\vspace{-10pt}
\end{figure}

\begin{figure*}[t]
{\centering{\includegraphics[width=1\linewidth]{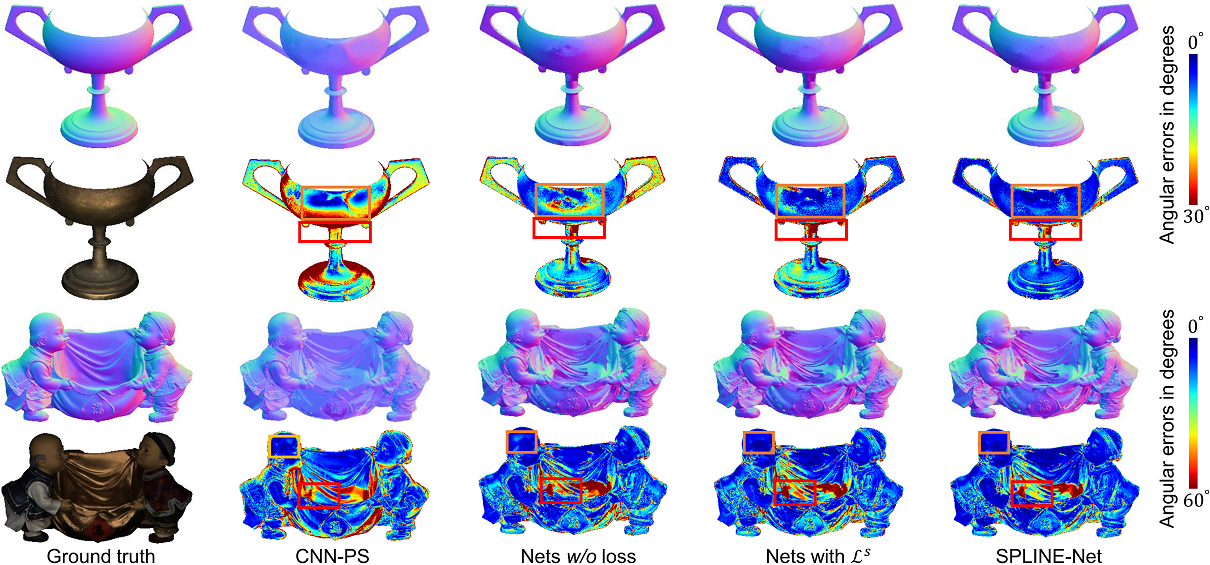}}
\vspace{-15pt}
\caption{Comparisons of normal maps and angular error maps (in degrees) on {\sc goblet} (top) and {\sc reading} (bottom) from {\sc DiLiGenT}~\cite{shi2019benchmark}.}
\label{fig:visual_ablation}}
\vspace{-3pt} 
\end{figure*}

\begin{figure*}[t]
{\centering{\includegraphics[width=1\linewidth]{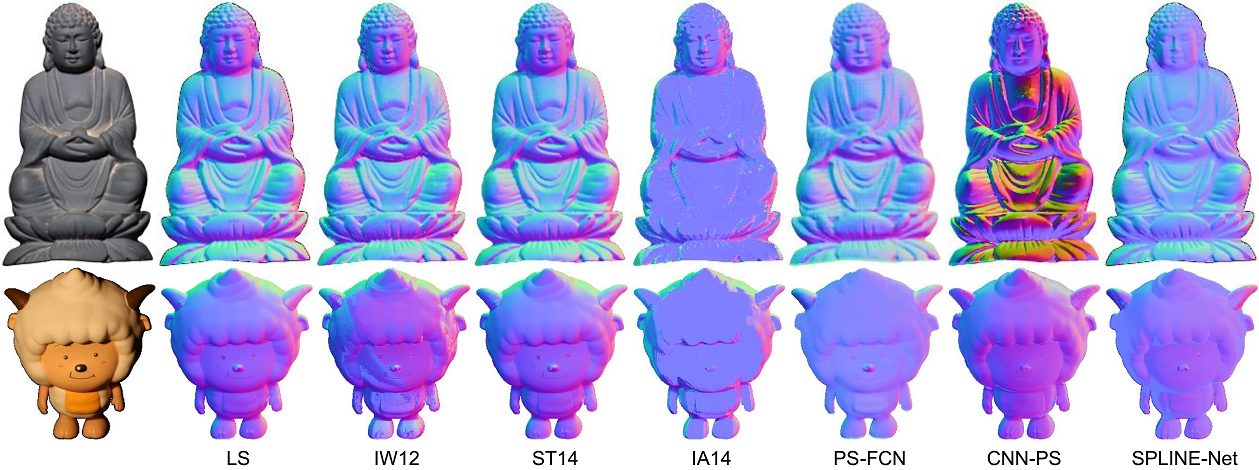}}
\vspace{-15pt}
\caption{Comparisons of normal maps on diffuse-dominant data {\sc buddha} (top) and {\sc sheep} (bottom).}
\label{fig:visual_limitation}}
\vspace{-8pt} 
\end{figure*}

The quantitative performance are reported in \Tref{tab:all_real}.
As can be observed, Nets \emph{w/o} loss significantly outperforms CNN-PS~\cite{ikehata2018cnn}, which verifies the effectiveness of using LI-Net to help the estimation of surface normals and prevent overfitting of directly regressing sparse observation maps to surface normals. 
The proposed symmetric loss and asymmetric loss help improve overall performance.

An illustration of observation maps generated by our method using different settings is displayed in \Fref{fig:maps}.
Our methods (Nets with $\mathcal{L}^s$ and SPLINE-Net) successfully inpaint regions damaged by global illumination effects and it tends to produce smooth ones even if the ground truth is not smooth.
Visual comparisons in \Fref{fig:visual_ablation} intuitively shows the advantages of each component of our method to facilitate accurate estimation of surface normals.

\section{Conclusion} 

This paper proposes SPLINE-Net to address the problem of photometric stereo with general reflectance and global illumination effects using a small number of images.
The basic idea of SPLINE-Net is to generate dense lighting observations from a sparse set of lights to guide the estimation of surface normals. 
The proposed SPLINE-Net is further constrained by the proposed symmetric and asymmetric loss functions to enforce isotropic constrain and perform outlier rejection of global illumination effects. 

\noindent\textbf{Limitations.}
Interestingly, even though deep learning based methods achieve superior performance for non-Lambertian reflectance, their performance drops for diffuse-dominant surfaces that can be well fitted by traditional methods with Lambertian assumption.
\Fref{fig:visual_limitation} illustrates results of four traditional methods and three deep learning based methods (including PS-FCN~\cite{chen2018ps}) on two real data\footnote{{\sc Buddha} is courtesy of Dan Goldman and Steven Seitz (found from \url{http://www.cs.washington.edu/education/courses/csep576/05wi/
/projects/project3}). {\sc Sheep} is from~\cite{shi2010self}.} with diffuse surfaces. 
Such results, which are consistent with those of {\sc ball} and {\sc pot1} in \Tref{tab:all_real}, indicate the limitation of deep learning methods for diffuse surfaces. 
To explicitly consider diffuse surface at the same time maintain the performance advantage on non-Lambertian surfaces for deep learning based methods can be one of further works.

\vspace{-1pt}
\section*{Acknowledgment}
\vspace{-1pt} 
This research was carried out at the Rapid-Rich Object Search (ROSE) Lab, Nanyang Technological University, Singapore.  
This work was supported in part by the National Research Foundation under NRF-NSFC Grant NRF2016NRF-NSFC001-098, the NTU's College of Engineering under Grant M4081746.D90, and the Infocomm Media Development Authority, Singapore.
This work was also supported in part by the National Natural Science Foundation of China under Grant 61872012, Grant 61661146005, and Grant U1611461.

{\small
\bibliographystyle{ieee_fullname}
\bibliography{egbib}
}

\end{document}